\PassOptionsToPackage{numbers, compress}{natbib}
\documentclass{article}

\usepackage[preprint]{neurips_2026} 

\usepackage{filecontents}

\usepackage{graphicx,epsfig,xspace,url,xspace,color,multirow}

\usepackage{amsmath,endnotes}
\usepackage{xurl}
\usepackage{enumitem}
\usepackage[update,prepend]{epstopdf}
\usepackage{array} 
\usepackage{varwidth} 
\usepackage{flushend}

\usepackage[linesnumbered, vlined, ruled]{algorithm2e}
\usepackage{algpseudocode}
\usepackage{soul}
\usepackage{makecell}
\usepackage{tikz}

\usepackage{tablefootnote}
\usepackage[para,online,flushleft]{threeparttable}
\usepackage[normalem]{ulem}

\usepackage{amssymb}
\usepackage{pifont}
\usepackage{xcolor}
\usepackage{booktabs}
\usepackage{makecell}
\usepackage[english]{babel}
\usepackage{blindtext}
\usepackage{xspace}
\usepackage{amsmath}
\usepackage[normalem]{ulem}
\usepackage{caption}
\usepackage{graphicx}
\usepackage{booktabs}
\usepackage{arydshln}
\usepackage{xurl}
\usepackage{subcaption}
\usepackage{wrapfig}
\usepackage[toc,title]{appendix}
\usepackage{afterpage}
\usepackage{tikz}
\usepackage{enumitem}
\usepackage{stfloats}
\usepackage[linesnumbered,ruled]{algorithm2e}
\usepackage{caption}
\SetCommentSty{mycommfont}
\usepackage{comment}
\usepackage{multirow}
\usepackage{xcolor}
\usepackage{graphicx}

\usepackage[svgnames]{xcolor} 
\usepackage{tcolorbox}
\usepackage{pifont}      
\usepackage{amsmath}     
\usepackage{amsfonts}    
\usepackage{amssymb}     
\usepackage{ulem}
\usepackage{siunitx}
\newcommand{\mysection}[1]{\vspace{-.1in}\section{#1}\vspace{-.05in}}
\newcommand{\mysubsection}[1]{\vspace{-.06in}\subsection{#1}\vspace{-.06in}}

\newcommand{\etc}{\textit{etc.}\xspace}
\newcommand{\ie}{\textit{i.e.,}\xspace}
\newcommand{\eg}{\textit{e.g.,}\xspace}

\newcommand{\system}{$\textsc{CLAP}$\xspace}	
\newcommand{\BULLET}{\vspace{+.00in} \noindent $\bullet$ \hspace{+.00in}}

\usepackage[textsize=tiny,colorinlistoftodos,textwidth=20mm]{todonotes} 

\begin{document}


\title{CLAP: Contrastive Latent-space Prompt Optimization for End-to-end Autonomous Driving}

\author{
  Ruiyang Zhu \\
  University of Michigan \\
  \texttt{ryanzhu@umich.edu} \\
  \And
  Yuehan He \\
  University of Michigan \\
  \texttt{mperform@umich.edu} \\
  \And
  Boyuan Zheng \\
  University of Michigan \\
  \texttt{boyuann@umich.edu} \\
  \And
  Zesen Zhao \\
  University of Michigan \\
  \texttt{hymanzzs@umich.edu} \\
  \And
  Ahmad Chalhoub \\
  University of Michigan \\
  \texttt{aachalho@umich.edu} \\
  \And
  Qingzhao Zhang \\
  University of Arizona \\
  \texttt{qzzhang@arizona.edu} \\
  \And
  Z. Morley Mao \\
  University of Michigan \\
  \texttt{zmao@umich.edu} \\
}

\maketitle

\begin{abstract}

End-to-end autonomous driving systems powered by Vision-Language-Action (VLA) models achieve strong performance on common driving scenarios, yet remain brittle in rare but safety-critical long-tail situations such as active construction zones and complex yielding geometries. In this paper, we present a method that addresses the long-tail challenging scenes beyond data scaling and model training. 
We introduce CLAP (\textbf{C}ontrastive \textbf{La}tent-space \textbf{P}rompt optimization), a location-aware adaptation framework that augments a frozen VLA driving model with per-roadblock soft prompts, optimized from crowdsourced data and retrieved on demand via Vehicle-to-Everything (V2X) communication.
Our approach rests on two observations from VLAs' latent space: (i)~at the VLA's hidden-state layer, scenarios from the same roadblock cluster tightly and occupy compact regions of the latent space; and (ii)~within a single roadblock, long-tail and normal frames are heavily intermixed in the latent representation, making it difficult to improve one without disturbing the other. \textsc{CLAP} addresses this via a two-stage pipeline: supervised contrastive learning to discover a roadblock-specific hard-scene direction, followed by directionally regularized prompt optimization that selectively improves challenging frames while preserving normal frame performance. On the NAVSIM benchmark with various state-of-the-art VLA backbones, \textsc{CLAP} reduces challenging scenario planning error by 24\% with no regression on normal frames, significantly improving planning performance.

\end{abstract}
\mysection{Introduction}
\label{sec:intro}

The promise of end-to-end autonomous driving~(AD) lies in replacing hand-engineered perception-planning stacks~\cite{apollo, autoware} with unified models that jointly learn perception, prediction, and control from raw sensor streams~\cite{chitta2022transfuser, hu2023planning, yuan2024drama, liao2025diffusiondrive, xing2025goalflow}. Recent Vision-Language Models~(VLMs), pre-trained on Internet-scale vision-text corpora, bring additional benefits: they encode rich common-sense knowledge about traffic rules and safety margins that are difficult to express in purely geometric representations. Building on this foundation, recent work has extended VLMs into Vision-Language-Action~(VLA) planners with large-scale driving datasets~\cite{li2025recogdrive, li2025drivevla, wang2025alpamayo, li2026unidrivevla, zhou2025autovla}, in both pre-training and fine-tuning stages, achieving substantial gains on end-to-end planning benchmarks.

\begin{figure*}[t]
    \centering
    \begin{subfigure}{0.48\linewidth}
        \centering
        \includegraphics[width=\linewidth,height=2.9cm]{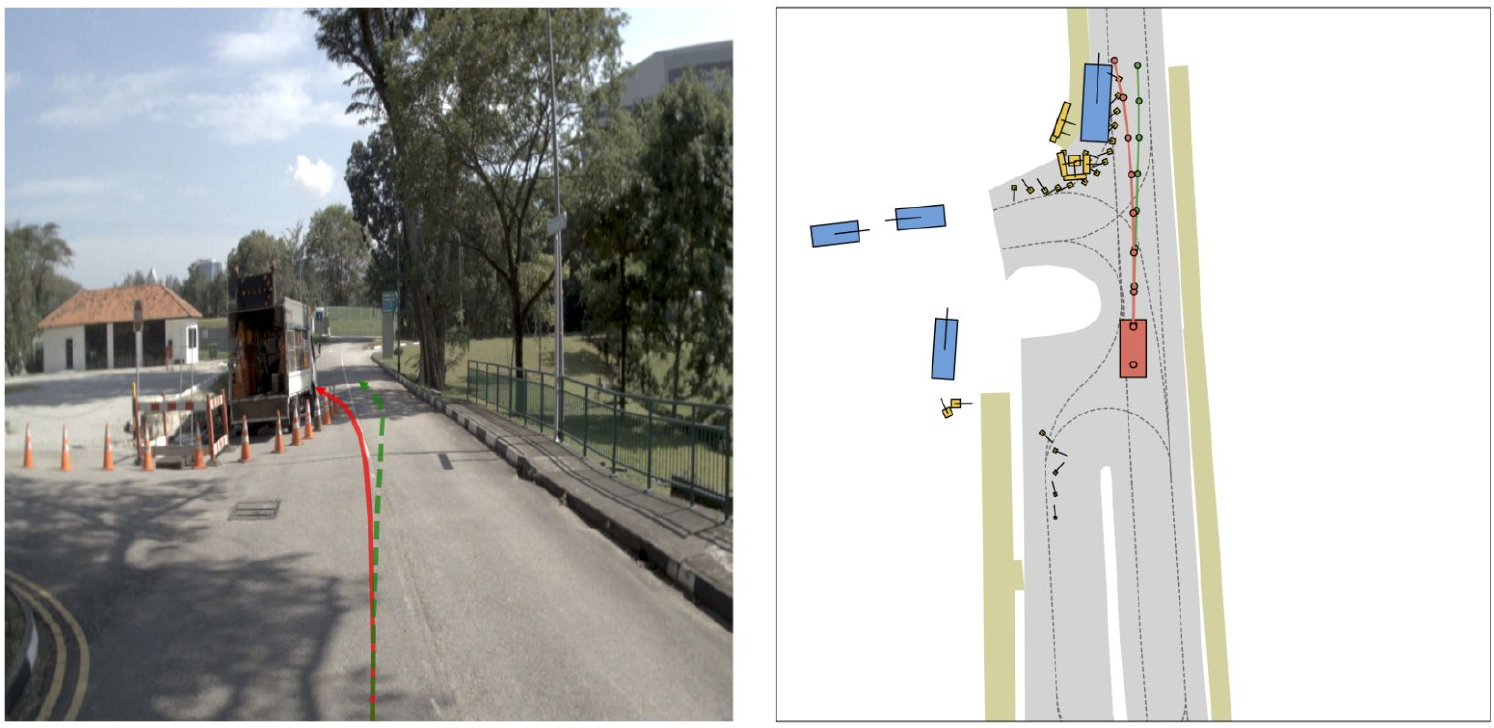}
        \caption{\footnotesize The model drive the vehicle to collide with a temporary workzone.}
        \label{fig:vla_robustness_1}
    \end{subfigure}%
    \hfill
    \begin{subfigure}{0.48\linewidth}
        \centering
        \includegraphics[width=\linewidth,height=2.9cm]{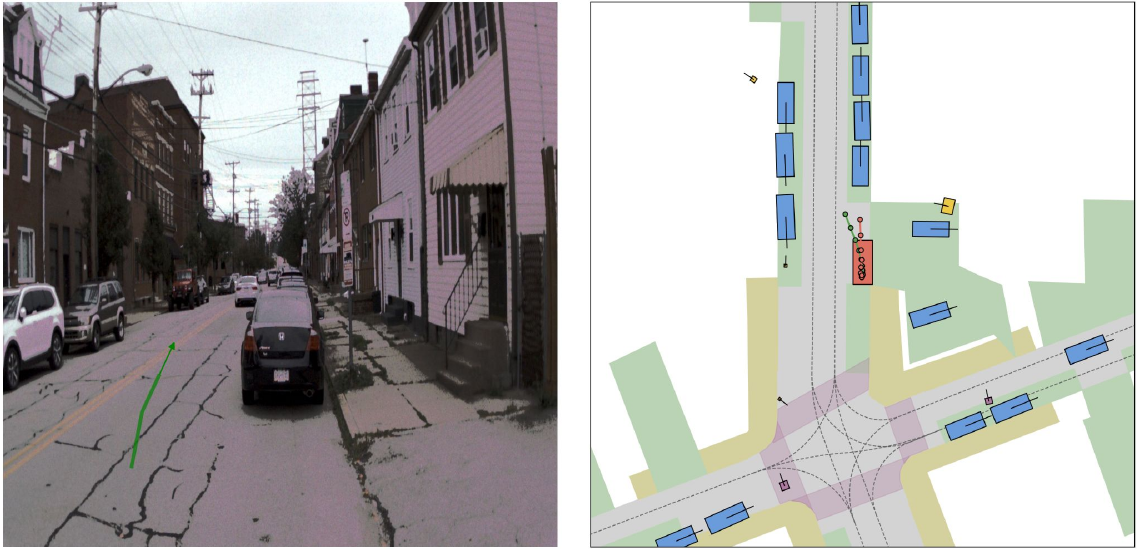}
        \caption{\footnotesize The model fails to detour when blocked by a parked vehicle.}
        \label{fig:vla_robustness_2}
    \end{subfigure}
    \vspace{-.1in}
    \caption{\small Examples of robustness issues of RecogDrive~\cite{li2025recogdrive} tested on NAVSIM~\cite{dauner2024navsim}. Red trajectories are produced by the driving model while green trajectories are human driver produced ground truth.}
    \vspace{-.2in}
\label{fig:vla_robustness}
\end{figure*}

Despite these advances, VLA-based AD systems share a well-known Achilles heel: \emph{long-tail scenario failures}. A \emph{long-tail scenario} is a driving situation that is (i)~often rare in the training distribution but (ii)~safety-critical in deployment. Canonical examples include active road construction with partial lane closures (Figure~\ref{fig:vla_robustness_1}), yielding under complex geometries and semantics (Figure~\ref{fig:vla_robustness_2}), and adverse weather combinations not well-represented in training. These cases cause state-of-the-art VLA planners to produce erratic trajectories or resort to overly conservative behaviors \cite{chen2024end, hu2025vision}.

The dominant approach to closing this gap is scaling data collection and fine-tuning the model, an axis along which recent VLA work continues to make substantial progress~\cite{li2025recogdrive, wang2025alpamayo, li2025drivevla, zhou2025autovla, xie2025s4, li2026unidrivevla}. These gains, however, become progressively harder to extend along the long-tail: rare events are sparse by construction~\cite{liu2024curse}, the combinatorial space of within-category variation (layouts, weather, time of day, surrounding traffic) makes exhaustive coverage intractable~\cite{bogdoll2022one, cheng2018quantitative}, and frequent fine-tuning on niche data risks catastrophic forgetting of previously learned competencies~\cite{luo2025empirical}.

\textbf{Key insight: long-tail failures can be patched locally.}
Rather than repairing a VLA planner for all possible long-tail scenarios, we observe that many failures are tied to specific road regions, such as recurring construction zones, complex intersections, or frequently blocked lanes. For such cases, the goal does not need to be category-level generalization across all construction zones or all blocked lanes. Instead, it is sufficient to improve behavior across repeated traversals of the same location.

\begin{figure*}[t]
    \centering
    \begin{subfigure}{0.33\linewidth}
        \centering
        \includegraphics[width=\linewidth]{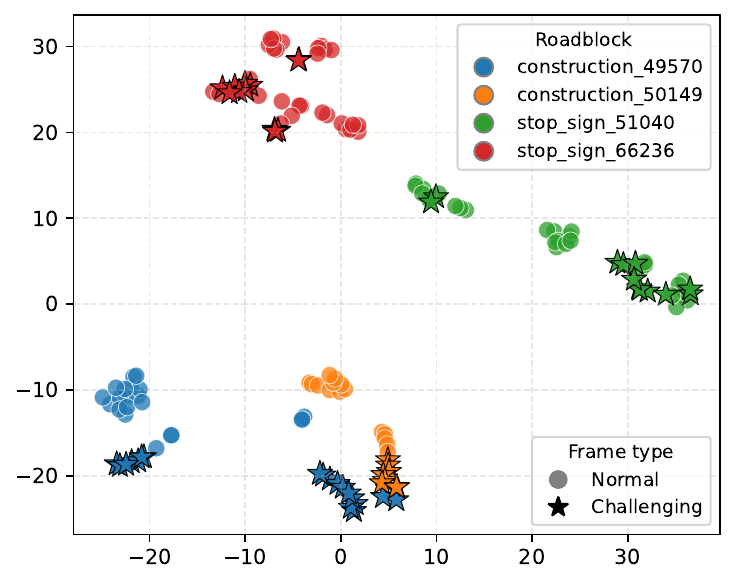}
        \vspace{-.15in}
        \caption{\footnotesize RecogDrive~\cite{li2025recogdrive}.}
        \label{fig:tsne}
    \end{subfigure}%
    \hfill
    \begin{subfigure}{0.33\linewidth}
        \centering
        \includegraphics[width=\linewidth]{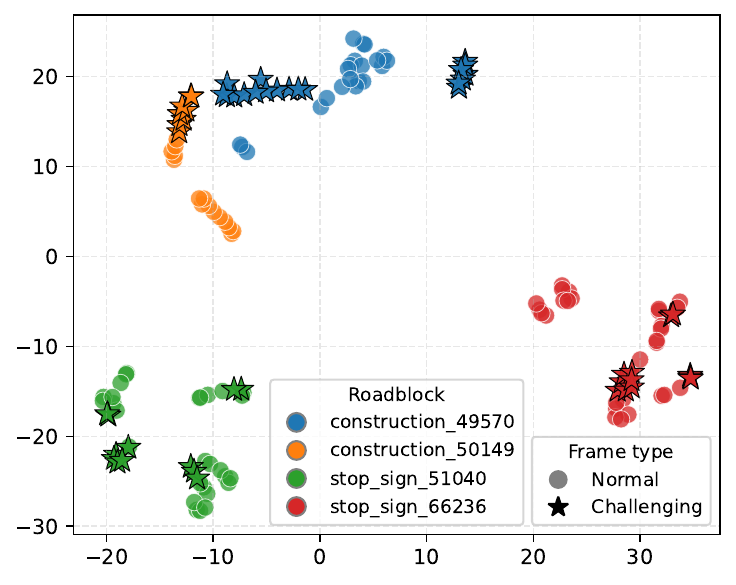}
        \vspace{-.15in}
        \caption{\footnotesize Alpamayo 1.5~\cite{wang2025alpamayo}.}
        \label{fig:tsne_alpamayo}
    \end{subfigure}
    \begin{subfigure}{0.33\linewidth}
        \centering
        \includegraphics[width=\linewidth]{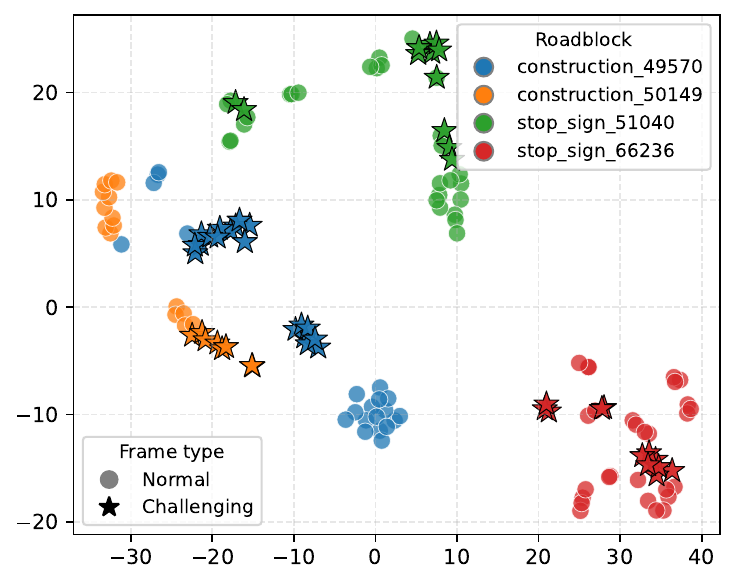}
        \vspace{-.15in}
        \caption{\footnotesize DriveVLA-W0~\cite{li2025drivevla}.}
        \label{fig:tsne_drivevla}
    \end{subfigure}
    \caption{\small 2D t-SNE~\cite{van2008visualizing} of hidden states from three state-of-the-art VLA planners on four NAVSIM regions. Across all three models, scenarios from the same region form clusters in the latent space.}
    \vspace{-.2in}
\label{fig:roadblock_tsne}
\end{figure*}

This locality is reflected in the model's latent space.
Figure~\ref{fig:roadblock_tsne} shows a t-SNE projection of three state-of-the-art VLA planners' (RecogDrive~\cite{li2025recogdrive}, Alpamayo~\cite{wang2025alpamayo}, and DriveVLA-W0~\cite{li2025drivevla}) hidden states over four spatial regions of the NAVSIM~\cite{dauner2024navsim} road topology: two construction zones and two stop-sign intersections. 
Across all three models, scenarios from the same region cluster tightly, reflecting shared visual structure (road geometry, landmarks, lane layout, \etc). On the other hand, different regions occupy distinguishable clusters rather than collapsing into a single category-level region (\eg the two stop-sign regions occupy distinct parts of the latent space). 
This suggests a natural unit of adaptation: we call each such region a \emph{roadblock}: a contiguous spatial region of the road whose traversals share a stable visual and geometric identity (same lane topology, dominant static obstacles and surrounding scene structure). A roadblock is an operational abstraction and can be instantiated in several ways, including (i)~HD-map segments \cite{wang2023openlane} bounded by change points in lane-topology, (ii)~geofenced polygons around persistent obstacles (\eg active construction), or (iii)~unsupervised clusters of GPS traces with similar visual embeddings \cite{berton2022rethinking}. 
This reframes the problem: rather than patching the model globally, a manifold too large for fine-tuning to cover, we can patch it \emph{per road}, where the manifold is small enough for a targeted intervention to cover cleanly.

\begin{figure}[t]
  \centering
  \includegraphics[width=\linewidth]{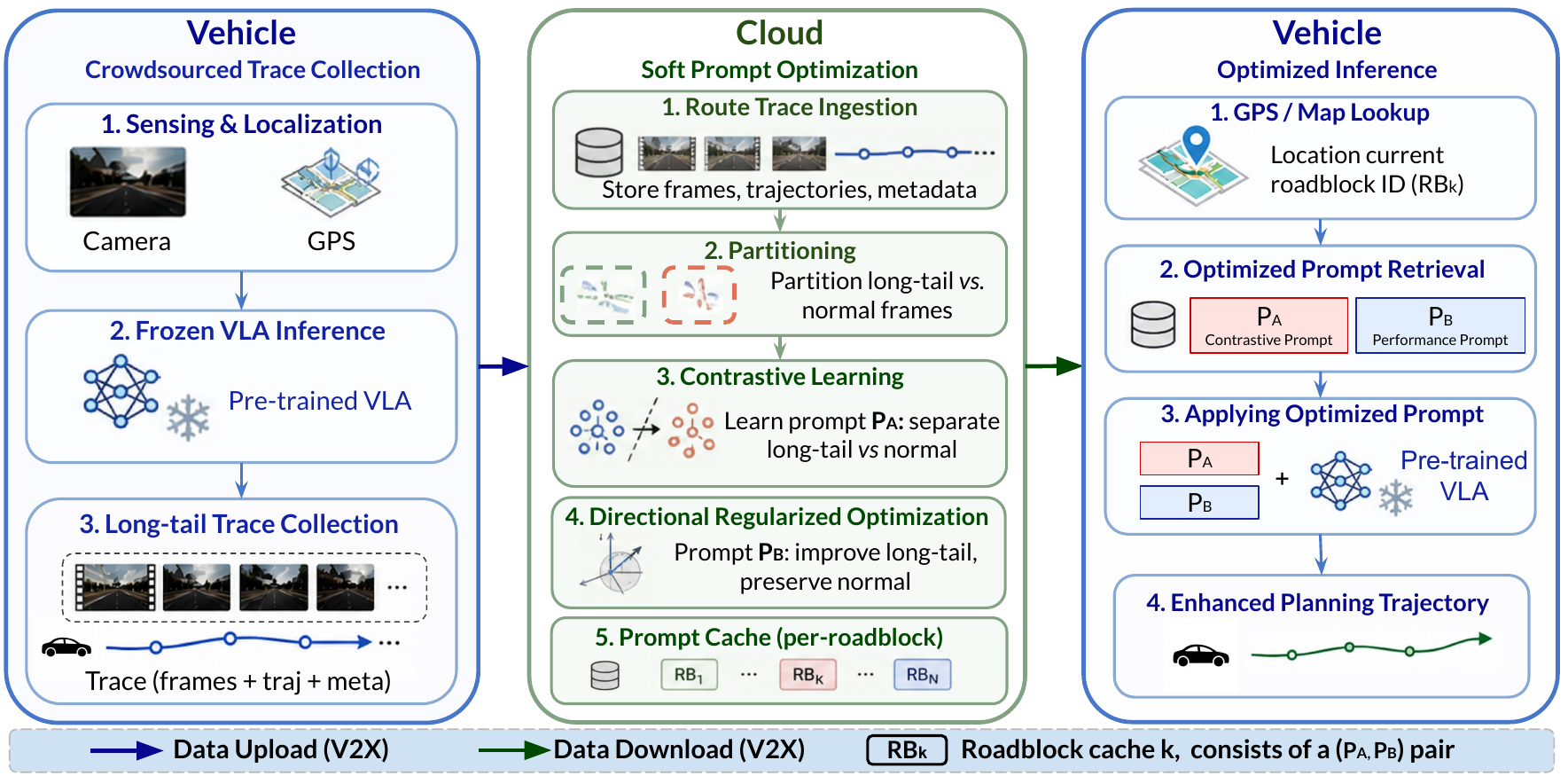}
  \caption{Overview of CLAP. Connected vehicles upload challenging traces over V2X; the cloud optimizes a per-roadblock prompt pair $(P_A, P_B)$; other vehicles retrieve and apply them to the frozen VLA when entering the same location.}
  \label{fig:arch}
  \vspace{-.2in}
\end{figure}

\textbf{Key method: soft prompt optimization through contrastive latent-space learning.}
We adopt soft-prompt tuning~\cite{lester2021power, wu2023infoprompt} as the adaptation mechanism, which optimizes continuous prompt embeddings while keeping the backbone model frozen.
This is particularly suitable for per-road adaptation: prompts can be learned, stored, and deployed independently across locations, avoiding interference between patches and eliminating the cost and risk of repeated fine-tuning. 
Soft-prompt tuning, however, brings its own challenge in this setting: a prompt optimized for challenging frames will also perturb normal frames on the same road, as challenging and normal frames are intermixed in latent space~\cite{makansi2021exposing} (\S\ref{sec:analysis:intermixed}). To address this, we design \textsc{CLAP}~(Figure~\ref{fig:arch}), a two-stage prompt optimization framework over a frozen VLA, instantiated \emph{per roadblock}:


\noindent\textbf{Stage~1 — direction discovery via supervised contrastive learning.}
We optimize an auxiliary prompt $P_A$ with a supervised contrastive (SupCon)~\cite{khosla2020supervised} objective that pushes hidden states of challenging frames away from those of normal frames. The principal direction of the separated representation yields $d^\star$, a compact descriptor of the \emph{challenging-scene direction}.

\noindent\textbf{Stage~2 — directionally regularized prompt optimization.}
We then optimize a performance prompt $P_B$ under a joint loss combining (a)~planning loss on challenging frames, 
(b) a directional regularizer steering $P_B$'s effect on challenging frames along $d^\star$,
and (c)~a normal-frame regularizer penalizing any shift in normal-frame representations. The resulting prompt improves long-tail behavior precisely where needed, with minimal collateral impact on common driving.

We envision \textsc{CLAP} deployed similarly to standard V2X communication-aided AD~\cite{li2021learning, wei2023asynchrony, zhu2024boosting, zhu2025scalable}: a central server aggregates challenging frames crowdsourced from connected vehicles (Figure~\ref{fig:arch}), optimizes per-roadblock prompts offline, and pushes the resulting prompts to vehicles on demand. 

We evaluate \textsc{CLAP} on challenging scenarios from NAVSIM~\cite{dauner2024navsim} benchmark with various SOTA VLA planners~\cite{li2025recogdrive, wang2025alpamayo, li2025drivevla},  demonstrating consistent improvements on long-tail scenarios without compromising normal driving performance.
The main contributions of the paper are:

\noindent\BULLET A per-road formulation of the long-tail problem: we show, via latent-space analysis across multiple SOTA VLAs, that the manifold of a single roadblock is closely clustered, motivating per-road rather than global optimization (\S\ref{sec:intro}).

\noindent\BULLET 
We propose \system framework to enhance long-tail scenario performance while preserving normal-frame planning, via a pipelined contrastive auxiliary prompt followed by a performance prompt optimized with directional and normal-frame regularizations (\S\ref{sec:design:clap}).

\noindent\BULLET We show \system reduce 24\% Average Displacement Error (ADE) on various VLA planners on NAVSIM challenging segments, without normal-frame regression (\S\ref{sec:evaluation}), achieving new state-of-the-art performance.






\mysection{Background and Related Work}
\label{sec:background}

\mysubsection{End-to-End VLM and VLA for Autonomous Driving}
\label{sec:bg:e2e}

Early end-to-end driving systems used convolutional or transformer architectures trained to regress waypoints from camera images, implicitly coupling perception and planning~\cite{chitta2022transfuser, hu2023planning, yuan2024drama, liao2025diffusiondrive, jia2025drivetransformer}. A newer generation integrates Large Language Models~(LLMs) or Vision-Language Models (VLMs) to bring language understanding and reasoning directly into the driving loop.
LMDrive~\cite{shao2024lmdrive} encodes multi-camera views and partial LiDAR point clouds via a vision encoder, then fine-tunes a frozen LLM plus a lightweight adaptor to produce waypoint outputs end-to-end given natural language driving instructions. EMMA~\cite{hwang2024emma}, developed by Waymo~\cite{waymo}, processes only camera images and text instructions and is built on Google's Gemini model~\cite{gemini}; it demonstrates that a single large generative model can handle joint perception, motion prediction, and planning. OpenEMMA~\cite{xing2025openemma} provides an open-source replication of this paradigm. ReCogDrive~\cite{li2025recogdrive} builds on the InternVL vision-language model~\cite{chen2024internvl, zhu2025internvl3} and is fine-tuned through reinforcement learning~\cite{sutton2018reinforcement}. AutoVLA~\cite{zhou2025autovla} similarly applies reinforcement fine-tuning, with adaptive reasoning tailored to driving. Alpamayo-R1~\cite{wang2025alpamayo} explicitly targets long-tail generalization by jointly training reasoning and action prediction, and DriveVLA-W0~\cite{li2025drivevla} introduces world-model amplification of the data-scaling law for VLA training. UniDriveVLA~\cite{li2026unidrivevla} unifies understanding, perception, and action planning within a single VLA backbone.
Despite their impressive performance on standard benchmarks, all these systems share a systematic weakness: performance degrades substantially on long-tail scenarios due to training distribution mismatch. Our work is orthogonal to the backbone design and can, in principle, be applied to any differentiable VLM-based planner to improve its long-tail performance.

\mysubsection{Prompt Optimization and Test-Time Adaptation}

\textbf{Hard prompt engineering} involves crafting natural language strings (\eg, ``Think step by step'') that improve zero-shot or few-shot performance~\cite{liu2023pre}. While effective for general tasks, the discrete search space makes systematic optimization difficult. \textbf{Soft prompts} replace human-readable tokens with continuous, learnable embeddings prepended to the input sequence. PrefixTuning~\cite{li2021prefix} and PromptTuning~\cite{lester2021power} showed that optimizing a small number of embedding parameters while freezing the backbone can match full fine-tuning on many NLP benchmarks. 
In the adversarial settings, GCG~\cite{zou2023universal} demonstrated that gradient-guided search over token space can reliably produce adversarial suffixes that jailbreak LLMs. AutoDAN~\cite{liu2023autodan} showed that the same principle applies in the multimodal domain. We adapt the optimization philosophy for a \emph{constructive} purpose: rather than eliciting harmful outputs, we steer the model toward safer, more appropriate driving decisions.
Test-time training~\cite{TTT} proposes updating a subset of model parameters at inference time using self-supervised auxiliary losses computed on the test input. TeST~\cite{TeST} extends this idea to semantic segmentation.
Our approach differs in two important ways: (1)~we do not update any model parameters, the prompt patch is fixed during retrieval and injection; and (2)~the optimization happens entirely in the cloud on a pre-collected set of crowdsourced data, making the online cost minimal.

\mysection{Motivating Experiment: Per-Roadblock vs. Shared Prompt Optimization} 
\label{sec:analysis}

As discussed in \S\ref{sec:intro}, our latent-space analysis (Figure~\ref{fig:roadblock_tsne}) revealed a locality property: scenarios from the same roadblock cluster tightly in the VLA's hidden-state space, while different roadblocks occupy geometrically distinct regions. Building on this, we test whether per-roadblock optimization outperforms a single shared optimization across roadblocks, and uncover a second geometric observation about within-roadblock structure that further shapes CLAP's design.

\textbf{Baseline: Soft prompt optimization.}
We apply the \emph{soft prompt optimization} method~\cite{lester2021power}: a sequence of $L$ continuous embeddings $P = [p_1, \dots, p_L] \in \mathbb{R}^{L \times d}$ injected after the visual tokens at the LLM input and optimized by gradient descent on the trajectory loss (MSE loss) via backpropagation. This recipe is standard in parameter-efficient adaptation~\cite{li2021prefix, lester2021power, zheng2024prompt} and serves as the motivating baseline in this section before \textsc{CLAP}'s contrastive two-stage refinement (\S\ref{sec:design}).

\begin{figure}[t]
  \centering
  \begin{minipage}{0.48\linewidth}
  \includegraphics[width=\linewidth]{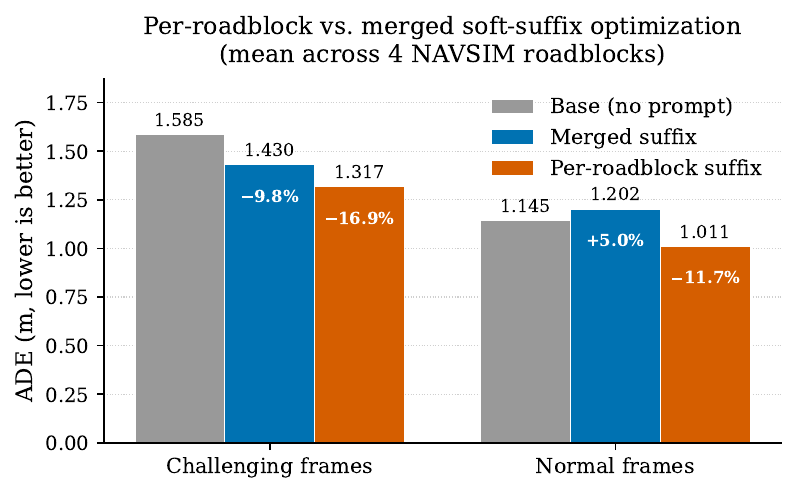}
  \caption{\small Per-roadblock vs. merged soft prompt optimization on RecogDrive, averaged across four NAVSIM roadblocks. Per-roadblock optimization yields larger reductions in both challenging-frame ADE while matching merged on normal-frame ADE.}
  \label{fig:scope_motivation}
  \end{minipage}\hfill
  \begin{minipage}{0.48\linewidth}
  \includegraphics[width=\linewidth]{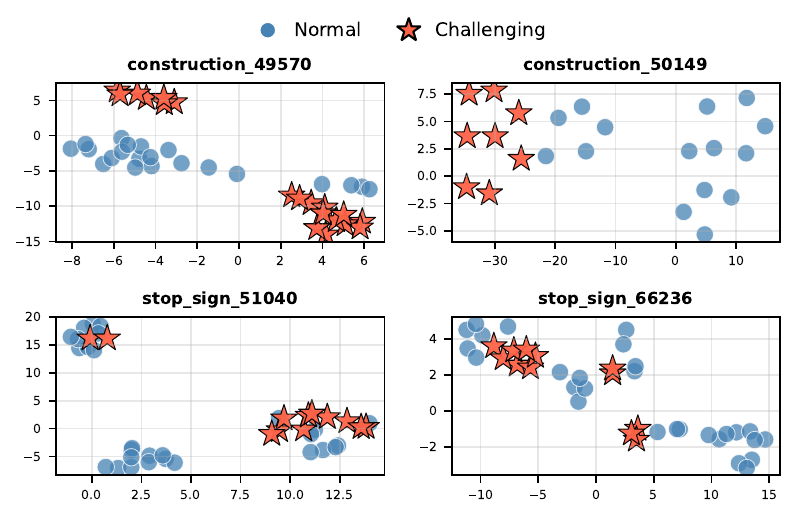}
  \caption{\small Per-roadblock t-SNE plots for challenging and normal frames within each roadblock. The challenging data and normal data frames are largely intermixed for most cases and cannot be easily separated.}
  \label{fig:per_rb_tsne}   
  \end{minipage}
  \vspace{-.15in}
\end{figure}

\mysubsection{Observation 1: per-roadblock scoping outperforms a shared prompt}
\label{sec:analysis:scope}

\textbf{Setup.}
We compare two scopes of soft prompt optimization against the RecogDrive~\cite{li2025recogdrive} model. Both methods use prompts of identical length and minimize the same trajectory loss on challenging frames; the only variable is the scope of optimization. \textbf{Merged}: a single prompt is optimized on the challenging frames of all four roadblocks and evaluated on each. \textbf{Per-roadblock}: four separate prompts are optimized, each on the challenging frames of one roadblock and evaluated on the same. 

Figure~\ref{fig:scope_motivation} reports challenging- and normal-frame ADE averaged across the four roadblocks.
The frozen backbone has challenging-frame ADE of $1.585$\,m. 
Merged-scope optimization reduces this to $1.430$\,m (\(-9.8\%\)), and per-roadblock optimization to $1.317$\,m (\(-16.9\%\)). The merged prompt underperforms despite being trained on four times more challenging frames than each per-roadblock prompt; the bottleneck is not data, it is the geometric mismatch between a single prompt and four distinct hidden-state manifolds.
The normal-frame results are more informative. Per-roadblock optimization improves normal-frame ADE by \(11.7\%\) given that challenging and normal frames at the same roadblock share visual structure and benefit from related representational adjustments. Merged optimization, in contrast, degrades normal-frame ADE by \(5.0\%\). 

\mysubsection{Observation 2: hard and normal frames are intermixed within a roadblock}
\label{sec:analysis:intermixed}

A naive view would expect challenging frames at a roadblock to occupy a distinct sub-region of the latent space relative to normal frames at the same roadblock. However, Figure~\ref{fig:per_rb_tsne} shows this is not the case. For each of the four roadblocks, we project the VLA's LLM-layer-0 hidden states without additional prompt hints via t-SNE, with challenging and normal frames coloured separately. Challenging frames do not form a clean island; they are heavily \emph{intermixed} with normal frames within each roadblock, occupying overlapping or interleaved regions rather than a separate cluster. 

This intermixing has a direct consequence for unconstrained prompt optimization. The trajectory-loss gradient on challenging frames points in the direction that locally reduces error on those frames, but because their hidden states are interleaved with normal frames', the same direction also perturbs normal-frame representations. Whether that perturbation helps or hurts on a given roadblock depends on the local geometry, which is why we observed mixed normal-frame outcomes in Observation~1: per-roadblock optimization happened to help normal frames on these four roadblocks, but it provides no mechanism to protect its performance. A roadblock with less favorable hard/normal overlap could see normal-frame regression even under per-roadblock scoping.

\textbf{Takeaway.}
The two observations together motivate the rest of \system's design. (a)~Scoping optimization to a single roadblock yields larger average challenging-frame gains than a shared optimization across roadblocks. (b)~Within a roadblock, optimization should explicitly account for not degrading normal-frame performance when focusing on improving challenging scenarios.
\mysection{Methodology}
\label{sec:design}

\mysubsection{Identifying Challenging Frames per Roadblock}
\label{sec:design:hardframes}

The goal of improving challenging-frame performance without degrading normal frames requires partitioning each roadblock's frames into challenging and normal subsets.
The partition is an input to \system: a deployed system could derive these labels from many signals (V2X-reported near-misses, ride-quality complaints, \etc), and the choice of signal is orthogonal to the prompt optimization framework. 
In this paper, we propose a viable approach via a VLM-based pipeline that serves both as our evaluation tool and as one partitioning component in deployed \system:

\textbf{Input.} The pipeline operates on crowdsourced route traces collected by \system's cloud optimizer (Figure~\ref{fig:arch}). Each trace consists of a sequence of camera frames, per-frame ego pose (position, speed, heading), the VLA-planned trajectory, and a ground-truth trajectory (\eg human-driven or annotated). The two trajectories together enable per-frame difficulty assessment, detailed below.

\textbf{(1) Semantic identification:} A strong VLM (Claude Sonnet 4.6~\cite{sonnet}) is given the front-camera frame sequence, per-frame ego speed and heading change, and brief map context for each route, and identifies route-level temporal segments involving non-trivial driving decisions. \textbf{(2) Empirical validation:} A semantically identified segment is retained only if its planning error under the frozen backbone exceeds the route mean by a margin $\delta$ (0.5\,m in our experiments); within retained segments, other frames are partitioned as normal. This ensures retained frames are challenging in both senses—semantically meaningful and empirically difficult for the planner. \textbf{(3) Spatial coherence:} Confirmed challenging frames should form spatially contiguous clusters in map coordinates, guarding against incidentally but geographically scattered outlier frames. Each surviving cluster receives a fine-grained label (e.g., \texttt{speed\_adapt\_\_nudge\_out\_\_odd\_construction} for a construction zone requiring an out-of-lane maneuver) used by Stage 1's contrastive weighting (\S\ref{sec:design:clap}). Full partitioning taxonomies and configurations are detailed in Appendix~\ref{app:hardframes}.

\mysubsection{CLAP: Two-Stage Contrastive Latent-Space Prompt Optimization}
\label{sec:design:clap}

\paragraph{Notations.}
Let $f_\theta$ denote a frozen VLA planner that predicts a future trajectory $\hat y = f_\theta(x, u)$ from camera observations $x$ and instruction tokens $u$. Throughout the process, $\theta$ is held fixed; CLAP only optimizes a pair of soft prompts injected at the LLM input. For a roadblock $\mathcal{R}$, let $\mathcal{D}_c^\mathcal{R}$ denote the challenging frames identified by the pipeline of \S\ref{sec:design:hardframes} and $\mathcal{D}_n^\mathcal{R}$ the remaining normal frames. CLAP optimizes one prompt pair $(\mathbf{P}_A^\mathcal{R}, \mathbf{P}_B^\mathcal{R})$ per roadblock, using only frames from $\mathcal{R}$.

The two prompts are continuous embedding sequences in the LLM input space, $\mathbf{P}_A^\mathcal{R} \in \mathbb{R}^{m \times d_{\text{llm}}}$ and $\mathbf{P}_B^\mathcal{R} \in \mathbb{R}^{n \times d_{\text{llm}}}$, spliced between the visual tokens and the instruction tokens:
\[
\underbrace{[\,v_1, \dots, v_K\,]}_{\text{visual}}\;\Vert\;
\underbrace{[\,\mathbf{P}_A^\mathcal{R}\,]}_{\text{Stage 1 trains}}\;\Vert\;
\underbrace{[\,\mathbf{P}_B^\mathcal{R}\,]}_{\text{Stage 2 trains}}\;\Vert\;
\underbrace{[\,u_1, \dots, u_L\,]}_{\text{instruction}}.
\]
Splicing here, rather than as a trailing suffix, ensures the prompts are causally visible to every instruction and trajectory token under the LLM's auto-regressive mask. 

\textbf{Stage 1 — contrastive direction discovery.}
Stage 1 trains $\mathbf{P}_A^\mathcal{R}$ with a single objective: induce a separation between challenging and normal frames in the planner's hidden-state geometry. 
For each frame $i$ we extract a representation $h_i = Pool\!\big(H^{(\ell)}_i(\mathbf{P}_A^\mathcal{R})\big) \in \mathbb{R}^{d_{\text{llm}}}$, where $H^{(\ell)}_i(\mathbf{P}_A^\mathcal{R})$ is the layer-$\ell$ hidden state of the LLM with $\mathbf{P}_A^\mathcal{R}$ present in the input, and $\mathrm{Pool}$ averages over the visual-token positions only ($\mathbf{P}_A^\mathcal{R}$ and instruction positions are excluded). $h_i$ is L2-normalized before the loss is applied. We use $\ell{=}0$ throughout the main experiments as we empirically find it effective.

Within a batch $\mathcal{B}$ containing both challenging and normal frames, we minimize the supervised contrastive (SupCon) loss~\cite{khosla2020supervised}, treating every challenging frame as an anchor:
\[
\mathcal{L}_{\text{SupCon}}
= -\frac{1}{|\mathcal{A}_c|}\!\sum_{i\in \mathcal{A}_c}\!
\frac{1}{\sum_p w_{ip}} \sum_{p \in \mathcal{P}(i)} w_{ip}\,
\log\!\frac{\exp\!\big(h_i \!\cdot\! h_p / \tau\big)}
            {\sum_{k \in \mathcal{B}, k \ne i} \exp\!\big(h_i \!\cdot\! h_k / \tau\big)},
\]
where $\mathcal{A}_c$ is the set of challenging frames in the batch, $\mathcal{P}(i)$ is the set of other challenging frames serving as positives, 
$w_{ip}=1$ if $i$ and $p$ share a fine-grained label.
The contrast set in the denominator includes both challenging and normal frames.
After Stage 1, we estimate the roadblock's \emph{hard-scene direction} $\mathbf{d}^\star$ as the top-$k$ right-singular vector of the residualized challenging representations $\{h_i - \bar h_n\}_{i \in \mathcal{D}_c^\mathcal{R}}$, where $\bar h_n$ is the mean normal-frame representation under $\mathbf{P}_A^\mathcal{R}$. Intuitively, $\mathbf{d}^\star$ captures the dominant axis along which Stage 1 has separated hard from normal in this specific roadblock. Stage 2 will steer the planning prompt along this axis.

\textbf{Stage 2 — directionally regularized prompt optimization.}
Stage 2 freezes $\mathbf{P}_A^\mathcal{R}$ and trains the deployable prompt
$\mathbf{P}_B^\mathcal{R}$ under a joint loss, with each term addressing complementary requirements:
\[
\mathcal{L}_{\text{total}}
= \mathcal{L}_{\text{plan}}
+ \lambda_1 \mathcal{L}_{\text{dir}}
+ \lambda_2 \mathcal{L}_{\text{reg}}.
\]

\noindent\textbf{(a) Planning loss.} The trajectory regression loss
(ADE) computed only on challenging frames:
\[
\mathcal{L}_{\text{plan}}(\mathbf{P}_B^\mathcal{R})
= \frac{1}{|\mathcal{D}_c^\mathcal{R}|}\!
\sum_{i \in \mathcal{D}_c^\mathcal{R}}
\mathrm{ADE}\ \!\big(\hat y_i,\, y_i^{\text{gt}}\big).
\]

\noindent\textbf{(b) Directional regularizer along $\mathbf{d}^\star$.}
This term encourages $\mathbf{P}_B^\mathcal{R}$ to displace hard-frame representations along the axis Stage 1 identified:
\[
\mathcal{L}_{\text{dir}}(\mathbf{P}_B^\mathcal{R})
= -\frac{1}{|\mathcal{D}_c^\mathcal{R}|}
\sum_{i \in \mathcal{D}_c^\mathcal{R}}
\big\langle\,
h_i^{(\ell)}(\mathbf{P}_A^\mathcal{R}, \mathbf{P}_B^\mathcal{R}) - h_i^{(\ell)}(\mathbf{P}_A^\mathcal{R}),\;
\mathbf{d}^\star
\,\big\rangle.
\]
By construction, $\mathbf{d}^\star$ is the axis along which Stage 1
optimized hard/normal separability; biasing $\mathbf{P}_B^\mathcal{R}$'s
update toward this axis prevents the planning gradient from drifting into
arbitrary directions that could affect normal-frame representations.

\noindent\textbf{(c) Normal-frame protection.}
This term explicitly penalizes any displacement of normal-frame representations relative to the Stage 1 baseline:
\[
\mathcal{L}_{\text{reg}}(\mathbf{P}_B^\mathcal{R})
= \frac{1}{|\mathcal{D}_n^\mathcal{R}|}
\sum_{j \in \mathcal{D}_n^\mathcal{R}}
\big\| h_j^{(\ell)}(\mathbf{P}_A^\mathcal{R}, \mathbf{P}_B^\mathcal{R}) - h_j^{(\ell)}(\mathbf{P}_A^\mathcal{R})\big\|^2.
\]
Even if $\lambda_1 \mathcal{L}_{\text{dir}}$ pushes challenging frames strongly along $\mathbf{d}^\star$, $\mathcal{L}_{\text{reg}}$ structurally regularizes $\mathbf{P}_B^\mathcal{R}$ from carrying a component that affects normal frames. This is the operationalization of the selectivity property that distinguishes CLAP from unconstrained prompt optimization (\S\ref{sec:analysis:scope}).

\textbf{Putting the two stages together.}
The two stages form a constructive chain. Stage 1 discovers the roadblock-specific direction $\mathbf{d}^\star$ along which challenging and normal frames are separable. Stage 2 exploits that direction: $\mathcal{L}_{\text{dir}}$ steers planning improvement along $\mathbf{d}^\star$, and $\mathcal{L}_{\text{reg}}$ protects the normal frame performance. Combining them together, the result is an optimized prompt whose support is geometrically restricted to the subspace where it can plausibly help and structurally prevented from leaking into the subspace where it could hurt regular scenario performance.

\mysection{Experimental Evaluation}
\label{sec:evaluation}

\mysubsection{Experiment Setup}
\textbf{Dataset and base VLA models.}
We evaluate on the NAVSIM benchmark~\cite{dauner2024navsim}, specifically on a subset of 42 roadblock sequences identified as challenging (construction zones, complex stop sign intersections, \etc) from \S\ref{sec:design:hardframes}. We sampled 3 routes (camera and pose grouped by roadblock ID and time, each spanning 6--7 seconds) for each roadblock from the trainval split for prompt optimization and use the test split for evaluation; per-roadblock prompts are trained only on training frames from each roadblock and evaluated on test frames from the same roadblock. In total, this yields 1{,}440 trainval frames and 3{,}810 test frames. To verify the effectiveness of CLAP, we adopt it across 3 SOTA VLA models for autonomous driving: ReCogDrive~\cite{li2025recogdrive}, Alpamayo-R1.5~\cite{wang2025alpamayo}, and DriveVLA-W0~\cite{li2025drivevla} to show that our design can be applied to different VLA architectures.

\textbf{Implementation details.}
We use soft prompts of length $m = $ 50 for $P_A$ and $n = $ 50 for $P_B$, both initialized from $\mathcal{N}(0, 1)$. Stage~1 trains $P_A$ for 50 epochs with $k=3$ and Stage~2 trains $P_B$ for 100 epochs at learning rate $1 \times 10^{-3}$ with Adam~\cite{kingma2014adam}, using $\lambda_1 = $ 0.1 and $\lambda_2 = $ 0.1.  All experiments run on $2 \times$ NVIDIA A40 GPUs~\cite{a40}.



\textbf{Baselines.}
We compare CLAP against: (1)~each VLA base model introduced above, (2)~\textbf{explicit notice}: a designed text hint prompt (e.g. ``Construction zone ahead. Reduce speed, 
prepare for lane change.'') generated by a strong VLM Claude Sonnet 4.6~\cite{sonnet}, and (3)~\textbf{unconstrained soft prompt:} a full-dimensional soft prompt optimized through backpropagation, following~\cite{lester2021power}.

\mysubsection{Main Results and Ablation Study}
\label{sec:eval:main}

\begin{table}[t]
\centering
\caption{\textbf{Main results: performance comparison across methods on NAVSIM challenging/normal splits.}
ADE@4s reported in meters; lower is better. Numbers in parentheses indicate
change relative to the off-the-shelf backbone (negative = improvement). \textit{Hard} denotes challenging frames at the evaluated roadblocks; \textit{Normal}
denotes normal frames at the same roadblocks; \textit{Overall} is the
frame-weighted average across both subsets (\ie testset ADE). Best result marked in
\textbf{bold}.}
\label{tab:main_results}
\setlength{\tabcolsep}{6pt}
\small
\begin{tabular}{l S[table-format=1.3] @{~} l  S[table-format=1.3] @{~} l  S[table-format=1.3] @{~} l}
\toprule
\textbf{Method} &
\multicolumn{2}{c}{\textbf{Hard ADE@4s $\downarrow$}} &
\multicolumn{2}{c}{\textbf{Normal ADE@4s $\downarrow$}} &
\multicolumn{2}{c}{\textbf{Overall ADE@4s $\downarrow$}} \\
\midrule
\textit{ReCogDrive (3 cameras)~\cite{li2025recogdrive}} & & & & & & \\
\midrule
ReCogDrive             & 2.125 &              & 1.007 &              & 1.562 &              \\
\quad + Explicit Notice         & 2.137 & ($+0.012$)   & 1.010 & ($+0.003$)   & 1.566 & ($+0.004$)   \\
\quad + Unconstrained Soft Prompt    & 1.835 & ($-0.290$)   & \bfseries 0.927 & \bfseries ($-0.080$) & 1.393 & ($-0.169$)   \\
\quad + \textbf{\textsc{CLAP} (Ours)} & \bfseries 1.426 & \bfseries ($-0.699$) & 0.934 & ($-0.073$) & \bfseries 1.227 & \bfseries ($-0.335$) \\
\midrule
\textit{Alpamayo-R1.5 (4 cameras)~\cite{wang2025alpamayo}} & & & & & & \\
\midrule
Alpamayo-R1.5         & 2.998 &              & 0.970 &              & 1.342 &              \\
\quad + Explicit Notice         & 3.130 & ($+0.132$)   & 1.091 & ($+0.121$)   & 1.467 & ($+0.125$)   \\
\quad + Unconstrained Soft Prompt    & 2.385 & ($-0.613$)   & 0.997 & ($+0.027$)   & 1.268 & ($-0.074$)   \\
\quad + \textbf{\textsc{CLAP} (Ours)} & \bfseries 2.281 & \bfseries ($-0.717$) & \bfseries 0.959 & \bfseries ($-0.011$) & \bfseries 1.185 & \bfseries ($-0.157$) \\
\midrule
\textit{DriveVLA-W0 (1 camera)~\cite{li2025drivevla}} & & & & & & \\
\midrule
DriveVLA-W0            & 2.116 &              & 1.011 &              & 1.243 &              \\
\quad + Explicit Notice         &  2.581  &    ($+0.465$)          &   1.409 &      ($+0.398$)         &   1.684 &     ($+0.441$)         \\
\quad + Unconstrained Soft Prompt    & 1.895 & ($-0.221$)   & 1.024 & ($+0.013$)   & 1.199 & ($-0.044$)   \\
\quad + \textbf{\textsc{CLAP} (Ours)} & \bfseries 1.804 & \bfseries ($-0.312$) & \bfseries 0.989 & \bfseries ($-0.022$) & \bfseries 1.146 & \bfseries ($-0.097$) \\
\bottomrule
\end{tabular}
\vspace{-.15in}
\end{table}

Table~\ref{tab:main_results} reports ADE@4s for all three backbones across all baselines. Three findings stand out:

\textit{Crafted natural-language explicit notice fails to improve frozen VLAs in challenging scenarios.} On Alpamayo-R1.5 and DriveVLA-W0, it even increases challenging-frame ADE by an average of 13\%. This suggests that crafted natural-language hints, even produced by a powerful model~\cite{sonnet}, are an unreliable mechanism for steering frozen VLA behavior in this setting.

\textit{Continuous soft prompt optimization, previously shown effective for NLP tasks~\cite{lester2021power, li2021prefix}, yields substantial challenging-frame gains but produces mixed normal-frame outcomes}: its performance regresses on Alpamayo-R1.5 and DriveVLA-W0. This matches the geometric concern raised in \S\ref{sec:analysis} (Observation~2): with hard and normal frames intermixed in the latent representation, an unconstrained prompt has no mechanism preventing its updates from influencing and potentially hurting normal driving performance.

\textit{\system reduces challenging-frame ADE by an average of 24\% across the three VLA models and is able to maintain the normal frame ADE.} On normal frames, \system even slightly improves performance on Alpamayo-R1.5 and DriveVLA-W0, and matches the unconstrained soft prompt baseline on ReCogDrive. This selectivity reflects \system's design goal: the directional regularizer $\mathcal{L}_{\text{dir}}$ confines updates to the contrastive direction $d^\star$, and the normal-frame regularizer $\mathcal{L}_{\text{reg}}$ structurally prevents leakage into the orthogonal complement. The improvements on challenging scenarios, along with the maintained performance on normal frames, make \system improve the overall ADE by an average of 14\%, demonstrating the effectiveness of our core idea: local prompt optimizations can efficiently improve VLA performance without finetuning the model.

\textbf{Impact of Contrastive Training.} We ablate \system's central design: Stage~1 contrastive direction discovery, using ReCogDrive as the backbone VLA. We compare RecogDrive with three configurations: \textbf{(i)} \textsc{CLAP} without Stage~1, where $P_B$ is optimized under $\mathcal{L}_{\text{plan}} + \lambda_2 \mathcal{L}_{\text{reg}}$ ($\mathcal{L}_{\text{dir}}$ omitted as $d^\star$ is unavailable); \textbf{(ii)} \textsc{CLAP} with $d^\star$ replaced by a random direction; and \textbf{(iii)} the full \textsc{CLAP} pipeline. As shown in Table~\ref{tab:ablation_stage1}, removing Stage~1 entirely leaves $\mathcal{L}_{\text{reg}}$ as the only constraint on $P_B$ and recovers a substantial portion of \textsc{CLAP}'s gain, indicating that normal-frame protection is a useful objective. Replacing $d^\star$ with a random direction, however, performs worse than omitting Stage~1 entirely. The full \textsc{CLAP} achieves the best result by combining two complementary ingredients: $\mathcal{L}_{\text{reg}}$ provides normal-frame protection, while $d^\star$ supplies the specific direction along which $P_B$ should optimize.

\begin{table}[t]
\centering
\caption{\textbf{Ablation: contribution of Stage~1 contrastive learning.} ADE@4s on ReCogDrive across the NAVSIM roadblocks. \textsc{CLAP} outperforms both ablated variants, confirming that the contrastive direction is not interchangeable with normal-frame regularization alone or with a random anchor.}
\label{tab:ablation_stage1}
\setlength{\tabcolsep}{8pt}
\small
\resizebox{\linewidth}{!}{%
\begin{tabular}{l S[table-format=1.3] S[table-format=1.3] S[table-format=1.3]}
\toprule
\textbf{Configuration} &
\textbf{Hard ADE@4s $\downarrow$} &
\textbf{Normal ADE@4s $\downarrow$} &
\textbf{Overall ADE@4s $\downarrow$} \\
\midrule
ReCogDrive                            & 2.125 & 1.007 & 1.562 \\
\textsc{CLAP} w/o Stage~1 ($\mathcal{L}_{\text{plan}} + \mathcal{L}_{\text{reg}}$ only) & {1.515} & {0.938} & {1.311} \\
\textsc{CLAP} w/ random $d^\star$                 & {1.647} & {0.999} & {1.341} \\
\textbf{\textsc{CLAP}}                     & \bfseries 1.426 & \bfseries 0.934 & \bfseries 1.227 \\
\bottomrule
\end{tabular}
}
\vspace{-.15in}
\end{table}

\begin{figure}[t]
  \centering
  \includegraphics[width=\linewidth]{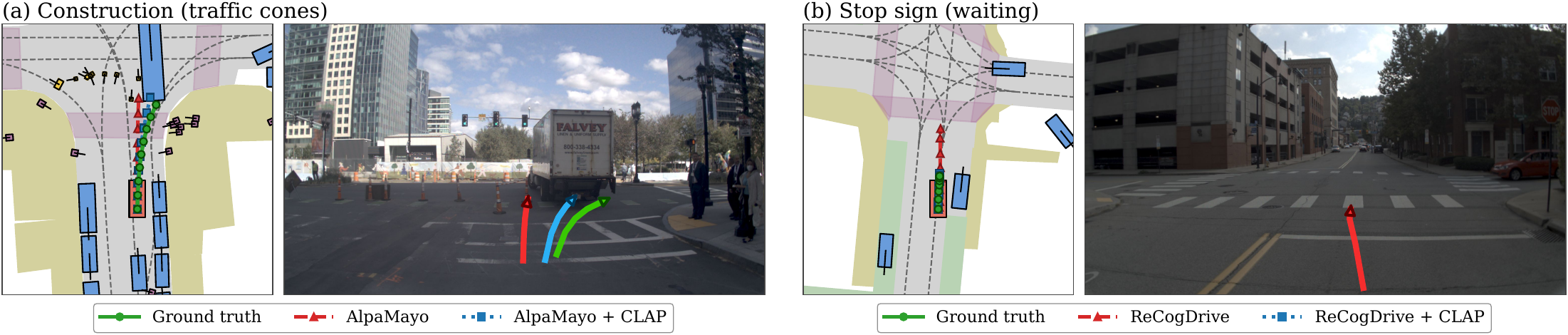}
  \caption{\small Planned trajectories on challenging frames before (red) and after (blue) \textsc{CLAP} adaptation. Vehicle with \textsc{CLAP} evades construction zone (left), and stops at stop sign (right)---no movement shown in camera.}
  \label{fig:trajectory_viz}
  \vspace{-.15in}
\end{figure}

\textbf{Behavioral trajectory comparison.} Figure~\ref{fig:trajectory_viz} compares predicted and ground-truth trajectories on two representative challenging scenarios. With \system, predicted trajectories track the human-driver ground truth more closely on a construction-zone bypass and a stop-sign yielding decision, with smoother dynamics, no stop-line violation, and no collision with static obstacles.

\mysubsection{Generalization Analysis through Data Augmentation}
\label{sec:eval:augmentation}

The limited challenging data from the training set after \S\ref{sec:design:hardframes} (5--20 frames per roadblock) inevitably reflects specific weather and lighting conditions, raising a deployment concern: a roadblock observed only on sunny afternoons may yield a prompt that fails to generalize when the same road is later traversed at night or in the rain~\cite{sakaridis2021acdc, zhao2025fishertune}.
To address this, we synthesize photorealistic visual variants using Qwen-Image-Edit~\cite{wu2025qwen}, a generative image-editing model, producing frames under weather and lighting conditions underrepresented in NAVSIM (Appendix~\ref{app:weather_aug}). The editor preserves scene geometry, agent positions, and lane structure, so the original ground-truth ego trajectory remains valid for each variant. 
We then evaluate CLAP on two additional test conditions:
(1) \textbf{Matched.}  Test frames from the same conditions as the challenging frames (identical to \S\ref{sec:eval:main}).
(2) \textbf{Shifted.} Frames re-rendered into rain and dusk, on test-split frames disjoint from training.

\begin{table}
\centering
\caption{\textbf{Generalization to condition shifts on ReCogDrive.} Hard- and normal-frame ADE@4s under matched conditions and shifted conditions (rendered into rain and dusk).}
\label{tab:augmentation}
\setlength{\tabcolsep}{6pt}
\small
\begin{tabular}{l S[table-format=1.3] S[table-format=1.3] S[table-format=1.3] S[table-format=1.3]}
\toprule
& \multicolumn{2}{c}{\textbf{Hard ADE@4s $\downarrow$}} & \multicolumn{2}{c}{\textbf{Normal ADE@4s $\downarrow$}} \\
\cmidrule(lr){2-3} \cmidrule(lr){4-5}
\textbf{Method} &
\textbf{Matched} & \textbf{Shifted} &
\textbf{Matched} & \textbf{Shifted} \\
\midrule
ReCogDrive               & 2.125 & 2.284 & 1.007 & 1.154 \\
\textsc{CLAP} w/o augmentation       & 1.426 & 1.877 & 0.934 & 1.178 \\
\textbf{\textsc{CLAP} + augmentation} & {\textbf{1.408}} & {\textbf{1.291}} & {\textbf{0.928}} & {\textbf{1.072}} \\
\bottomrule
\end{tabular}\
\vspace{-.15in}
\end{table}

Two findings stand out from Table~\ref{tab:augmentation}. First, augmentation improves performance even at matched conditions. \system~+~augmentation achieves the lowest hard-frame ADE, despite not including testset augmented variants in testing.
Second, augmentation is what enables \textsc{CLAP}'s gains to transfer across conditions. 
Without augmentation, hard-frame ADE rises by 32\% under condition shift; with augmentation, it drops to 1.291 and also improves normal ADE by 7\%. For the V2X-aided deployment of \textsc{CLAP}, this means a roadblock's prompt, once optimized at the cloud from augmented sets of crowdsourced challenging frames, can be deployed across various conditions at the same location.

\mysection{Discussion and Limitations}
\label{sec:discussion}

\textbf{Scope of generalization.}
Cross-weather generalization beyond the conditions tested in \S\ref{sec:eval:augmentation} remains an open question. The generative augmentation pipeline is designed to address part of this gap, but real-world validation at scale, including more driving diversity and a broader set of adverse conditions~\cite{sakaridis2021acdc} is needed to provide better evidence for CLAP's generalizability.

\textbf{Scenario homogeneity within a roadblock.}
Stage 1's contrastive objective treats all challenging frames within a roadblock as a single positive class. This implicitly assumes each roadblock contains one dominant challenging scenario type; when multiple distinct difficulty modes coexist (\emph{e.g.}, a construction zone and a partially occluded stop sign together), the contrastive signal mixes heterogeneous failure modes and weakens the recovered direction $d^\star$. How to delineate optimization regions so that each contains a single homogeneous challenge type, \eg by subdividing roadblocks by challenge type or by recovering multiple $d^\star$ axes per region, remains an open design question.

\mysection{Conclusion}
\label{sec:conclusion}

We present CLAP, a contrastive latent prompt optimization framework for end-to-end VLA autonomous driving that targets long-tail scenarios without modifying backbone model weights. 
By combining contrastive learning with directional regularization in prompt optimization, \system achieves performance improvements on challenging driving situations while preserving base model competency on normal scenarios. Our results on NAVSIM demonstrate a 24\% ADE reduction on challenging driving scenarios where state-of-the-art VLAs produce high planning errors. We believe the prompt optimization paradigm represents a practically efficient way to improve VLA performance for autonomous driving beyond data scaling and model fine-tuning.
\bibliographystyle{abbrvnat}
\bibliography{acmart}

\newpage
\appendix

\mysection{Appendix}

\begin{figure}[!htbp]
  \centering
  \begin{subfigure}{\linewidth}
    \includegraphics[width=\linewidth]{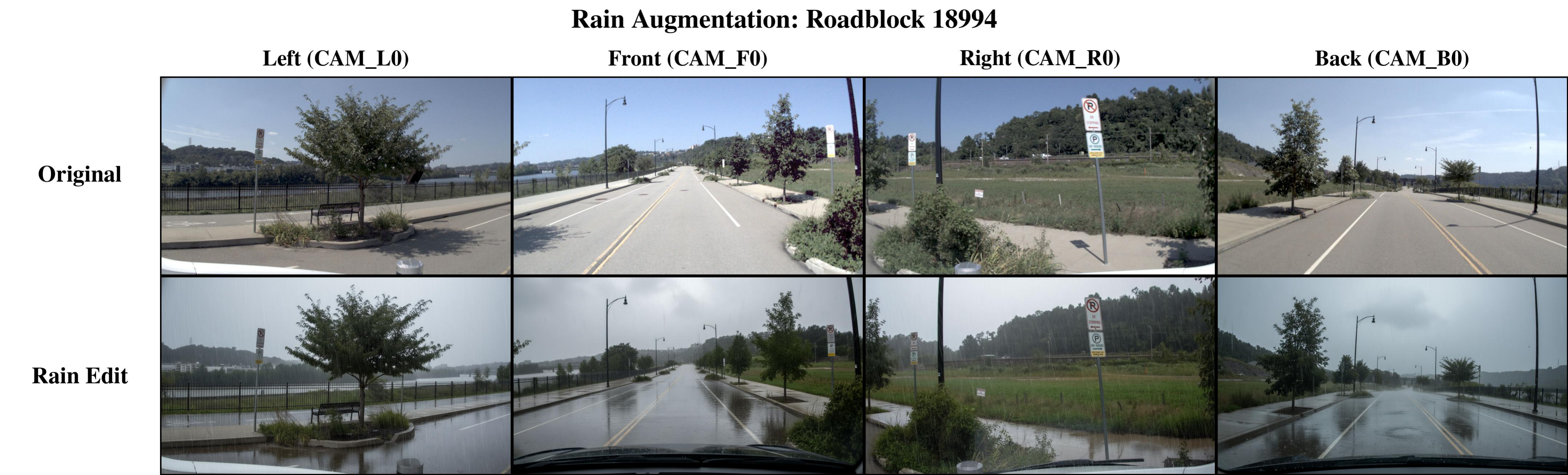}
    \label{fig:rain_aug_18994}
  \end{subfigure}\\[2pt]
  \begin{subfigure}{\linewidth}
    \includegraphics[width=\linewidth]{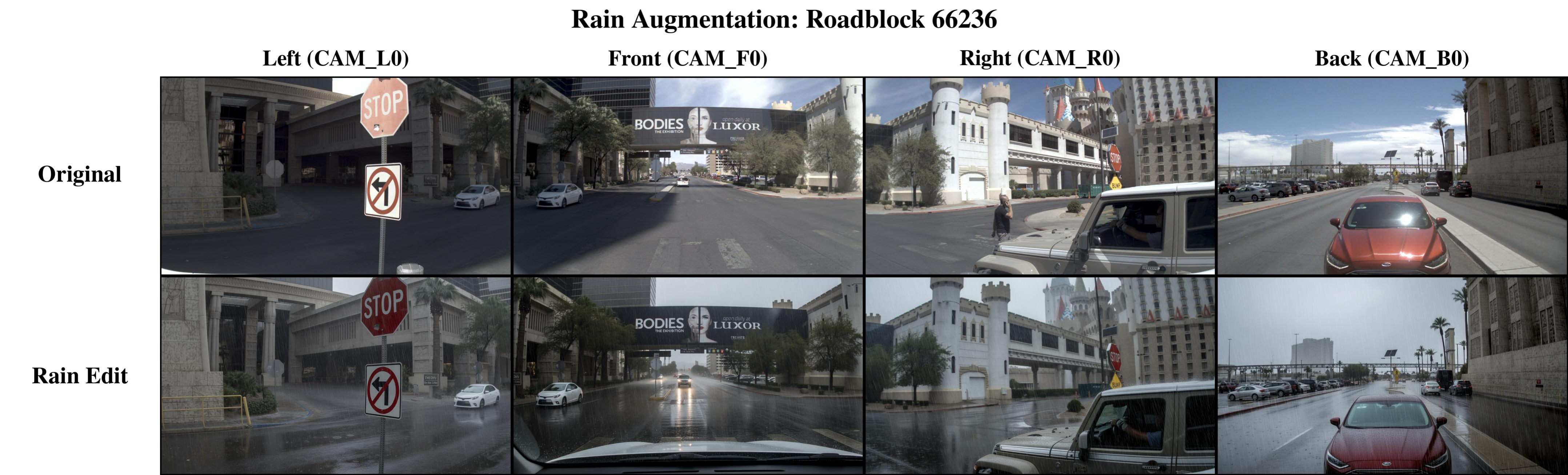}
    \label{fig:rain_aug_66236}
  \end{subfigure}\\[2pt]
  \begin{subfigure}{\linewidth}
    \includegraphics[width=\linewidth]{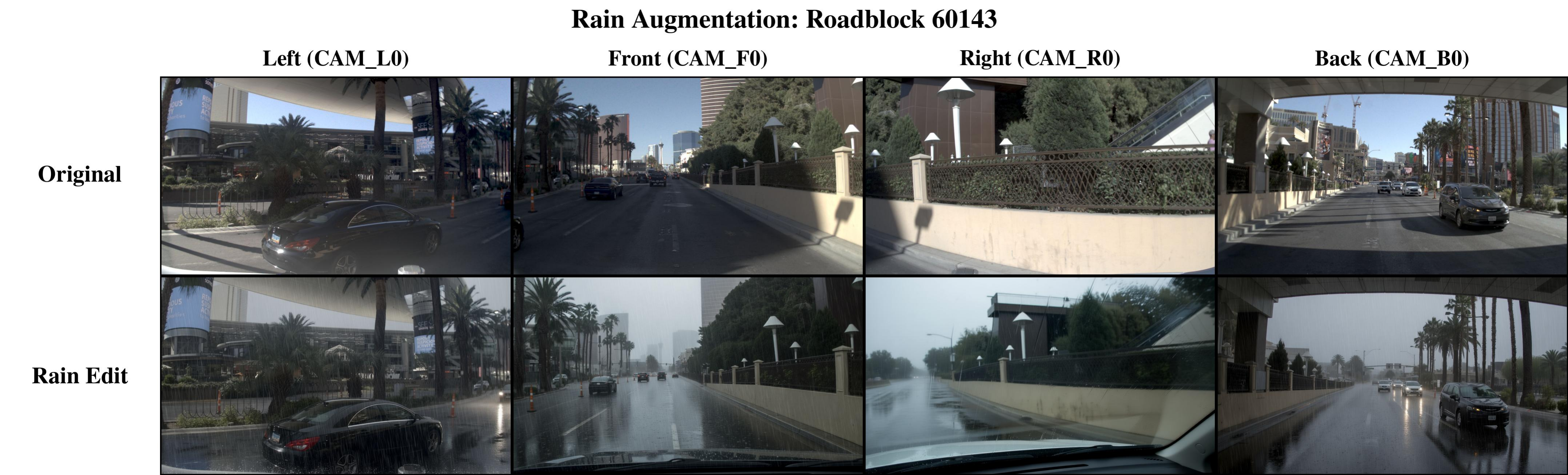}
    \label{fig:rain_aug_60143}
  \end{subfigure}
  \caption{\small Rain augmentation examples across three roadblocks. In each panel, columns are left, front, right, and back cameras; top row is the original NAVSIM frame, bottom row is the rain edit.}
  \label{fig:rain_aug}
  \vspace{-.1in}
\end{figure}

\subsection{Roadblock Challenging Frames and Normal Frame Data Construction} 
\label{app:hardframes}

CLAP requires a partition of each roadblock's frames into challenging and normal. We obtain this partition with a pipeline that combines an advanced vision-language model's semantic priors with empirical validation against the planner's own behavior. 

\textbf{Step 1 — Semantic identification.}
For each route through the roadblock, we provide a strong external VLM (Claude Sonnet 4.6~\cite{sonnet} in our implementation) with the front-camera frame sequence, ego speed and heading change per frame, and brief map context (intersection presence, lane count, road type). The model identifies temporal segments involving non-trivial driving decisions and emits a structured record per segment: a longitudinal decision label (\texttt{stop\_static}, \texttt{yield\_agent}, \texttt{speed\_adapt}, \dots), a lateral decision label (\texttt{lane\_keep}, \texttt{nudge\_out}, \texttt{turn}, \dots), the critical scene component (\texttt{yield\_stop\_high}, \texttt{critical\_obj\_high}, \texttt{odd\_construction}, \dots), and a short reasoning trace. Operating at the route level rather than frame-by-frame gives the VLM the temporal context required to identify where decision difficulty begins and ends.

\textbf{Step 2 — Empirical validation against planner behavior.} A semantically challenging segment is not necessarily challenging for the planner under analysis. We retain a Step-1 candidate segment only if its mean ADE under the frozen backbone exceeds the route mean by a margin $\delta$ ($\delta = 0.5$\,m in our experiments):
\[
\overline{\mathrm{ADE}}_{\text{segment}}
> \overline{\mathrm{ADE}}_{\text{route}} + \delta.
\]
Within retained segments, individual frames whose ADE is below the route mean are reclassified as normal. This two-filter rule ensures that a frame is challenging in both senses: semantically meaningful and empirically difficult for the specific planner being patched.

\textbf{Step 3 — Spatial coherence.}
Confirmed challenging frames within a roadblock should form one or a
small number of spatially contiguous clusters. If they originate from
multiple routes, we verify that the clusters lie in close geographic
proximity in $(x,y)$ map coordinates. This guards against treating an
incidentally hard but geographically scattered group of frames as a
single roadblock-specific phenomenon.

\textbf{Step 4 — Fine-grained taxonomy labels.}
After three steps, each surviving challenging cluster receives a fine-grained label of the form
\[
\texttt{\{long\_decision\}\_\_\{lat\_decision\}\_\_\{component\}\_\_\{uncertainty\}},
\]
e.g., 
\texttt{speed\_adapt\_\_nudge\_out\_\_odd\_construction}
(construction zone requiring an out-of-lane maneuver). 

\begin{figure}[!htbp]
  \centering
  \begin{subfigure}{\linewidth}
    \includegraphics[width=\linewidth]{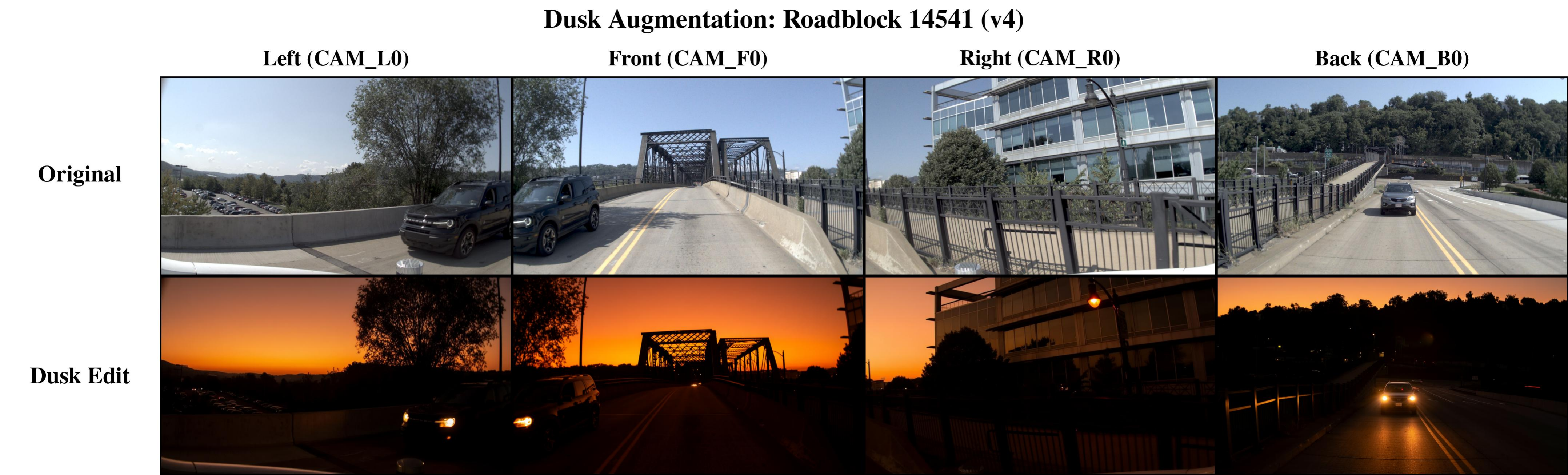}
    \label{fig:dusk_aug_14541}
  \end{subfigure}\\[2pt]
  \begin{subfigure}{\linewidth}
    \includegraphics[width=\linewidth]{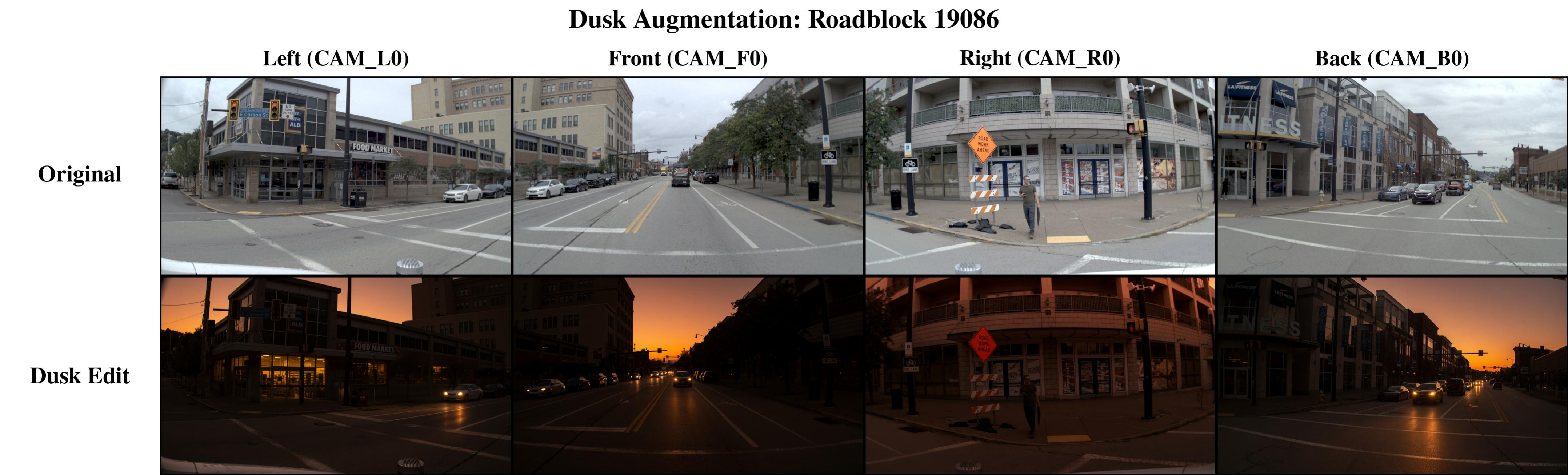}
    \label{fig:dusk_aug_19086}
  \end{subfigure}\\[2pt]
  \begin{subfigure}{\linewidth}
    \includegraphics[width=\linewidth]{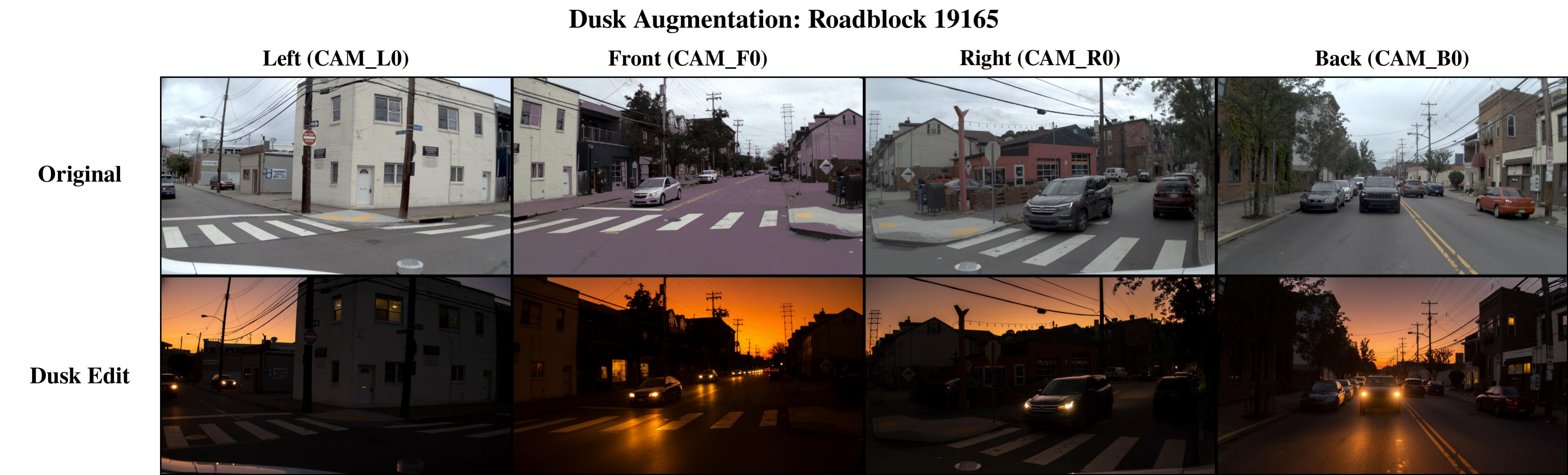}
    \label{fig:dusk_aug_19165}
  \end{subfigure}
  \caption{\small Dusk augmentation examples across three roadblocks. In each panel, columns are left, front, right, and back cameras; top row is the original NAVSIM frame, bottom row is the dusk edit.}
  \label{fig:dusk_aug}
  \vspace{-.1in}
\end{figure}

\subsection{Data Augmentation}
\label{app:weather_aug}

The augmentation experiment in \S\ref{sec:eval:augmentation} evaluates whether a roadblock-level prompt remains effective when the same location is observed under weather and lighting conditions that are rare in the original NAVSIM logs. We generate condition-shifted frames with an image-to-image editing pipeline built on Qwen-Image-Edit~\cite{wu2025qwen}, treating augmentation as an appearance-only transformation: road geometry, lane topology, traffic participants, camera pose, and ground-truth trajectory labels are all inherited from the original NAVSIM frame.

\textbf{Setup.}
We apply rain augmentation to 42 roadblocks drawn from the NAVSIM trainval and test splits. For each roadblock, every frame recorded by all four onboard cameras (front CAM\_F0, left CAM\_L0, right CAM\_R0, back CAM\_B0) is edited. Frames from the same timestamp across cameras are assigned the same condition preset, ensuring multi-camera consistency: the same weather appears simultaneously in all views. Augmented frames from the trainval split are used during prompt optimization; augmented test-split frames that are held out during optimization are used exclusively for evaluation, so no edited frame seen at training time appears in the evaluation set.

\textbf{Editing prompts.} Below are the prompts used to generate rainy and dusk variations of the drive routes:
\begin{quote}
\textbf{Rain:} \textit{``Edit this front-facing driving scene into realistic rainy weather. Keep the camera viewpoint, road geometry, lane markings, traffic participants, and scene layout consistent with the original image. Add overcast sky, wet road, rain streaks, puddles, and reduced visibility while preserving the original dashcam scene structure.''}

\textbf{Dusk:} \textit{``Edit this front-facing driving scene into realistic dusk. Keep the camera viewpoint, road geometry, lane markings, and scene layout identical to the original image. Critically, every vehicle on the road must remain at the exact same position, with the same shape, size, color, and orientation as in the original — do not add, remove, move, or alter any car. Only the weather, lighting, and sky may change. Apply twilight atmosphere with warm orange sky, dimmer ambient light, slightly darker road surface, and realistic headlight visibility — without modifying any vehicle.''}
\end{quote}

\textbf{Output construction.}
Each edited frame is saved with the original frame token as its filename stem plus a \texttt{\_rain} suffix, so it can be joined to the original NAVSIM metadata without ambiguity. The ground-truth future trajectory, roadblock id, challenging/normal label, and fine-grained challenge label are reused from the corresponding original frame. In data augmentation experiments, we replace the original camera images with the edited ones and reuse the same metadata like vehicle pose in the experiment.

Figure~\ref{fig:rain_aug} show synchronized four-camera examples from three different roadblocks, illustrating that the edit preserves scene geometry and traffic participants while producing visually consistent rainy-weather appearance across all camera views.

We additionally apply dusk augmentation, simulating low-sun lighting conditions at the same roadblock locations. Figure~\ref{fig:dusk_aug} show four-camera dusk examples across three roadblocks.
\subsection{Experiment Details}
\label{appendix:exp_details}

This section expands the experimental setup of \S5.1 with details too
fine-grained for the main text but useful for reproducibility, including
per-roadblock data composition, backbone-specific representation
extraction, and the full training schedule for each stage.

\paragraph{Trace structure and route sampling.}
Each driving scenario in our experiments follows the NA VSIM
convention~\cite{dauner2024navsim}: a $2$\,s history window
($4$ frames at $2$\,Hz) plus a $4$\,s ground-truth future
($8$ waypoints at $2$\,Hz). The planner predicts the $8$-waypoint future
trajectory from the multi-camera observation at the current frame.
For each of the $42$ roadblocks evaluated in \S5, we sample $3$ routes
from the NA VSIM \emph{trainval} split, each spanning $6$--$7$\,s of
continuous driving (i.e.\ $12$--$14$ frames per route at $2$\,Hz), to
form the per-roadblock training pool. Test frames are drawn from the
NA VSIM \emph{test} split; the partitioning pipeline of
\S\ref{sec:design:hardframes} is applied separately to each split. Across
all 42 roadblocks, the training pool contains $1{,}440$ frames
($323$ challenging $+$ $1{,}117$ normal) and the test pool contains
$3{,}810$ frames ($1{,}664$ challenging $+$ $2{,}146$ normal).

\paragraph{Backbone-specific representation extraction.}
The Stage~1 representation $h_i = \mathrm{Pool}\!\big(H^{(\ell)}_i(\mathbf{P}_A)\big)$
(\S\ref{sec:design:clap}) requires a per-backbone definition of which positions
$\mathrm{Pool}$ averages over, since the three VLA architectures expose
visual content differently at the LLM input:
\begin{itemize}[leftmargin=1.3em,itemsep=2pt,topsep=2pt]
  \item \textbf{ReCogDrive}~\cite{li2025recogdrive} (InternVL, $d_{\text{llm}}{=}1{,}536$): visual content is
    represented by \texttt{<IMG\_CONTEXT>} sentinel tokens whose
    embeddings are replaced at runtime by ViT patch features via masked
    scatter. We pool over the LLM positions occupied by these sentinels.
  \item \textbf{Alpamayo-R1.5}~\cite{wang2025alpamayo} (Qwen3-VL,
    $d_{\text{llm}}{=}4{,}096$): visual content uses
    \texttt{<image\_pad>} sentinels with the same masked-scatter
    mechanism. We pool over those positions ("vis\_all" pool).
  \item \textbf{DriveVLA-W0}~\cite{li2025drivevla} (Emu3,
    $d_{\text{llm}}{=}4{,}096$): visual content is encoded as discrete
    VQ codes that live directly in the LLM vocabulary; no scatter is
    needed. We pool over the LLM positions occupied by VQ-code tokens.
\end{itemize}
We extract from layer~$\ell{=}0$ of the LLM for ReCogDrive and Alpamayo and use $\ell{=}14$ for DriveVLA, which we empirically found works well. All representations are L2-normalized before the SupCon loss is applied.

\begin{figure}[!htbp]
  \centering
  \includegraphics[width=\linewidth]{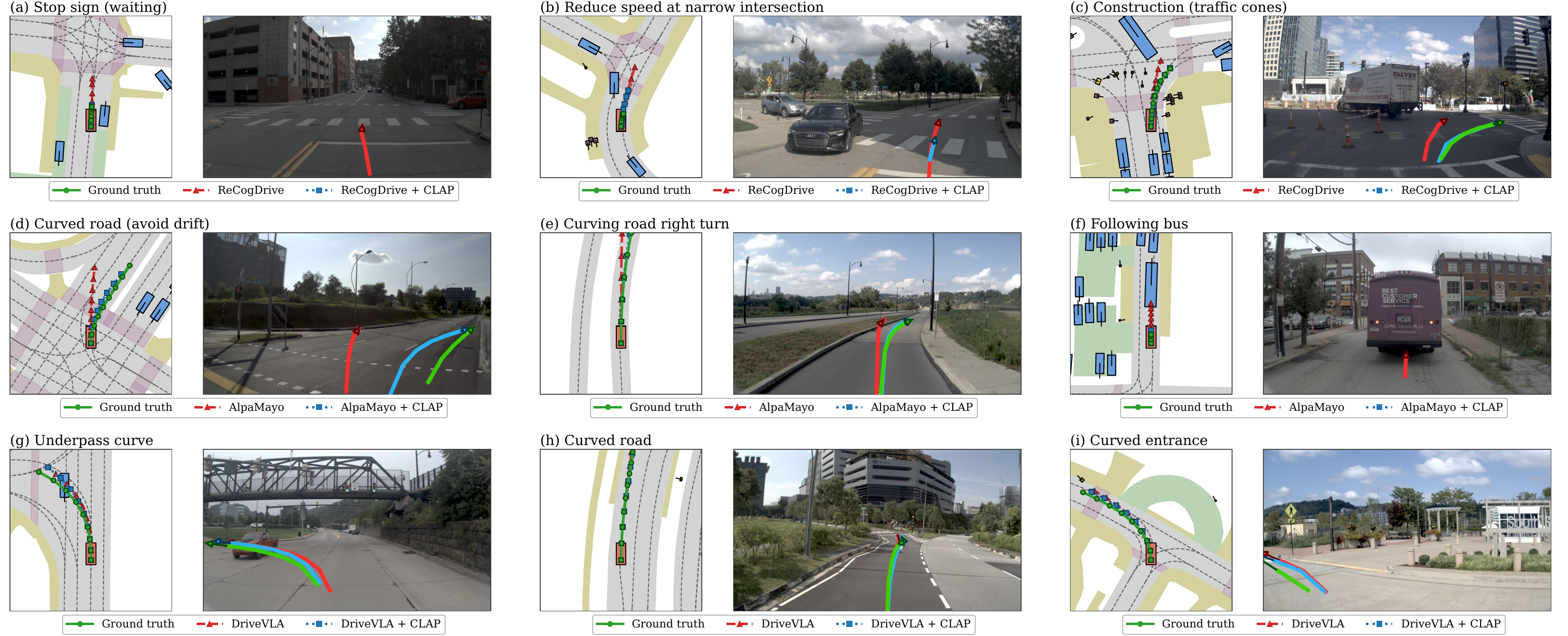}
  \caption{\small Predicted trajectories on additional challenging
    frames across three VLA backbones. Rows: ReCogDrive (a--c),
    Alpamayo-R1.5 (d--f), DriveVLA-W0 (g--i). Each panel pairs a
    bird's-eye-view map (left) with the front camera (right).
    Green: ground truth. Red: frozen backbone. Blue: backbone + CLAP.}
  \label{fig:traj_grid}
\end{figure}

\subsection{Additional Performance Visualization}
\label{app:additional_viz}

Figure~\ref{fig:traj_grid} extends the qualitative comparison of
Figure~\ref{fig:trajectory_viz} with additional challenging scenarios
per backbone, covering all three VLA planners evaluated in
\S\ref{sec:evaluation}. Each row corresponds to one backbone
(ReCogDrive top, Alpamayo-R1.5 middle, DriveVLA-W0 bottom), and each
column shows a different scenario. Within each panel, the left view
is the bird's-eye-view map with surrounding agents and the right view
is the front camera; green denotes the human-driver ground truth, red
the frozen backbone's prediction, and blue the same backbone with
CLAP applied.

Across all three backbones, the CLAP-adapted trajectories track the
ground truth more closely than the bare backbone in a range of
characteristic failure modes: stopping accurately at a stop sign
(a), reducing speed through a narrow intersection
(b), navigating around construction cones without colliding (c),
holding lane through curves rather than drifting straight (d, h),
executing right turns through curved geometry (e, i), and matching
the leading-vehicle pace when following a bus (f). The corrections
span scenario types and backbone architectures, indicating that the
per-roadblock prompts capture genuine challenges from roadblock envrionments and improve the planning decisions.

\end{document}